\documentclass{article}

    \usepackage[preprint]{neurips_2025}

\usepackage[utf8]{inputenc} 
\usepackage[T1]{fontenc}    
\usepackage{url}            
\usepackage{booktabs}       
\usepackage{amsfonts}       
\usepackage{nicefrac}       
\usepackage{microtype}      

\usepackage{colortbl}
\usepackage{multirow}

\usepackage{wrapfig} 
\usepackage{graphicx} 
\usepackage{array}    
\usepackage{tocloft}
\usepackage {titletoc}
\definecolor{iccvblue}{rgb}{0.21,0.49,0.74}

\usepackage{amssymb}
\usepackage[table,xcdraw]{xcolor}
\usepackage{wrapfig}
\usepackage{caption}
\usepackage{pifont}
\definecolor{citecolor}{HTML}{2980b9}
\definecolor{linkcolor}{HTML}{c0392b}
\definecolor{sem}{HTML}{2E75B6}
\definecolor{tok}{HTML}{F3B000}
\usepackage[hidelinks,breaklinks=true,colorlinks,citecolor=citecolor,linkcolor=linkcolor]{hyperref}

\usepackage{subcaption} 

\usepackage{amsmath} 

\newcommand{\cmark}{\ding{51}} 

\newcommand{\methodname}{\textsc{\textbf{LeanPO}}}

\title{\textit{\methodname{}}: Lean Preference Optimization for Likelihood Alignment in Video-LLMs}

\author{Xiaodong Wang$^{1,2}$ \quad Jinfa Huang$^{1}$ \quad Li Yuan$^{1,2}$  \quad Peixi Peng$^{1,2}$\thanks{Corresponding author.}   \\
 {$^{1}$Peking University \quad $^{2}$Peng Cheng Laboratory} \\
{\tt\small\{wangxiaodong21s@stu., pxpeng\}@pku.edu.cn}
\\
\\
Code: \textcolor{red}{\url{https://github.com/Wang-Xiaodong1899/LeanPO}}
}

\begin{document}

\maketitle

\vspace{-8mm}
\begin{abstract}
Most Video Large Language Models (Video-LLMs) adopt preference alignment techniques, e.g., DPO~\citep{rafailov2024dpo}, to optimize the reward margin between a winning response ($y_w$) and a losing response ($y_l$). However, the likelihood displacement observed in DPO indicates that both $\log \pi_\theta (y_w\mid x)$ and $\log \pi_\theta (y_l\mid x) $ often decrease during training, inadvertently boosting the probabilities of non-target responses. In this paper, we systematically revisit this phenomenon from LLMs to Video-LLMs, showing that it intensifies when dealing with the redundant complexity of video content. To alleviate the impact of this phenomenon, we propose \emph{Lean Preference Optimization} (\methodname{}), a reference-free approach that reformulates the implicit reward as the average likelihood of the response with respect to the policy model. A key component of \methodname{} is the reward-trustworthiness correlated self-generated preference data pipeline, which carefully infuses relevant prior knowledge into the model while continuously refining the preference data via self-reflection. This allows the policy model to obtain high-quality paired data and accurately estimate the newly defined reward, thus mitigating the unintended drop. In addition, we introduce a dynamic label smoothing strategy that mitigates the impact of noise in responses from diverse video content, preventing the model from overfitting to spurious details. Extensive experiments demonstrate that \methodname{} significantly enhances the performance of state-of-the-art Video-LLMs, consistently boosting baselines of varying capacities with minimal additional training overhead. Moreover, \methodname{} offers a simple yet effective solution for aligning Video-LLM preferences with human trustworthiness, paving the way toward reliable and efficient Video-LLMs. 
\end{abstract}
\section{Introduction}
\label{sec:intro}

Recent advanced Multi-modal Large Language Models (MLLMs) are built on powerful Large Language Models (LLMs) with visual encoders, capable of multi-modal understanding, reasoning, and interaction~\cite{sun2024generative,lin2024vila,liu2024improved,liu2024llavanext,zhang2024llavanextvideo,li2024llavaonevision,zhang2024llavavideo}. 
These models show remarkable improvement in images, videos, and other multi-modality tasks, continuously boosting the open-source MLLMs' development. 
However, since video content is more complex and diverse than images, Video-LLMs' responses may not be properly grounded in the video, which makes the development of Video-LLMs challenging.


\begin{figure*}
    \centering
    \includegraphics[width=\linewidth]{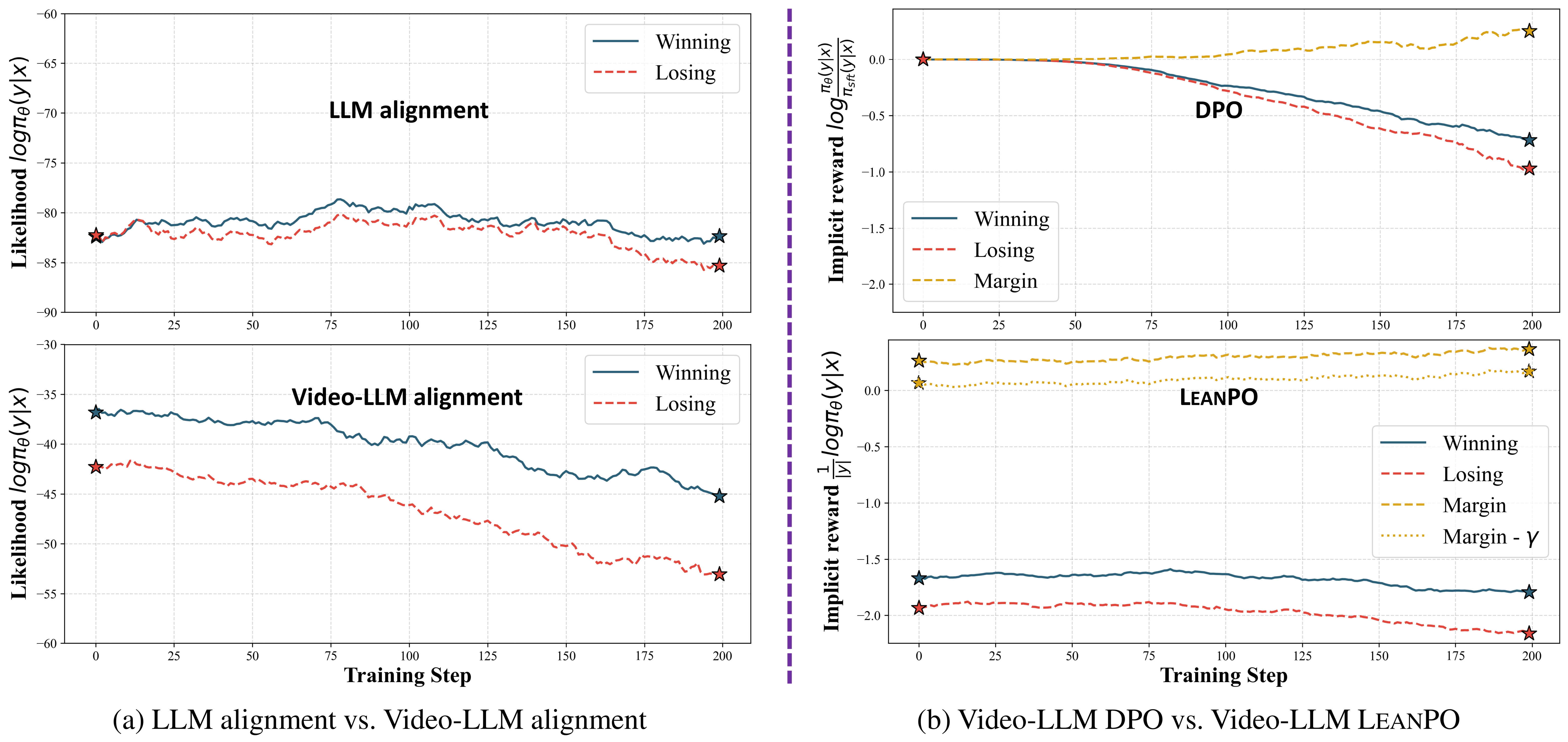}
    \vspace{-3mm}
    \caption{(a). Comparison between a LLM alignment based on Llama3-8B-Instruct~\cite{llama3modelcard} and a Video-LLM alignment based on LLaVA-NeXT-Video~\cite{zhang2024llavanextvideo} using DPO algorithm. (b). Comparison of the implicit reward between DPO and our proposed \methodname{} on a Video-LLM (LLaVA-NeXT-Video-7B). The proposed \methodname{} is immune to the likelihood displacement, making the reward in preference alignment training more stable and promoting efficient alignment.}
    \label{fig:motiv}
    \vspace{-5mm}
\end{figure*}



Due to the learned bias in Video-LLMs' large-scale pre-training and supervised fine-tuning (SFT), Video-LLMs often face misalignment and hallucinations when they are reasoning information from videos, which affects the trustworthiness in many practical applications~\cite{gunjal2024detecting}. 
Inspired by LLMs alignment works, most Video-LLMs alignment works adopt Direct Preference Optimization (DPO)~\cite{rafailov2024dpo} to align Video-LLMs with human values or intentions~\cite{vlm-rlaif,ahn2024isrt,zhang2024hound,li2025temporalPO}. DPO involves using human or AI's preference feedback for the model's responses, to collect preference pairs consisting of the winning response $y_w$ and the losing response $y_l$, and the loss is to maximize the margin between the reward of the winning and the losing responses for the policy model. However, both the likelihood of winning response \(\log \pi_\theta (y_w\mid x)\) and the losing response \(\log \pi_\theta (y_l\mid x)\) often decrease simultaneously during the DPO training. This phenomenon is widely known as the \emph{likelihood displacement} problem and has been discussed in many prior works on LLM alignment~\cite{yang2025dposhift,tajwar2024dpocapacity,razin2024unintentional,rafailov2024r}. This causes an unexpected increase in the probabilities of non-target responses. This phenomenon is attributed to various factors, such as model capacity~\cite{tajwar2024dpocapacity}, multi-token interference~\cite{pal2024smaug} or the sub-optimal training~\cite{rafailov2024r}. However, likelihood displacement phenomenon has not been explored for Video-LLMs.

In this paper, we systematically revisit likelihood displacement phenomenon from LLMs to Video-LLMs. First, we compare the likelihood curves of winning and losing responses between an LLM alignment based on Llama3-8B-Instruct~\cite{llama3modelcard} and a Video-LLM alignment based on LLaVA-NeXT-Video-7B~\cite{zhang2024llavanextvideo}, both trained using DPO algorithm, as shown in Fig.~\ref{fig:motiv}. (a). From LLMs to Video-LLMs, the likelihood of winning and losing responses drops more dramatically. 
Next, considering the reward definition in DPO, where the reward is given by $\log( \pi_\theta (y\mid x)/\pi_\texttt{sft}(y\mid x))$. We observe the corresponding implicit reward curve for Video-LLM DPO alignment in the upper part of Fig.~\ref{fig:motiv}. (b). Due to the more severe likelihood displacement phenomenon, the rewards for both responses also decline sharply. The winning reward should have remained the same or increased, but it actually decreased. This interplay between likelihood and reward affects the training dynamics, ultimately making the optimization of Video-LLMs inefficient. As previously discussed for LLMs~\cite{tajwar2024dpocapacity,pal2024smaug}, the exacerbation of this phenomenon in Video-LLMs during the DPO process may stem from the higher model capacity required for video understanding. 

Rewards are susceptible to the phenomenon of decreasing likelihood. Our intuition is that during training, the likelihood of a response to the policy model drops rapidly, which is too far from the likelihood of the response in the SFT model, making the ratio-type reward used by DPO unstable. Thus, to mitigate the negative effect of the likelihood displacement phenomenon, in this paper, we reformulate the implicit reward as the average likelihood of the response generated by the policy model, and such a definition removes the dependence on reference models. The reward is defined as $\frac{1}{|y|}\log \pi_{\theta}(y)$, and the reward curves of winning responses become more stable in the training, as shown in the bottom of Fig.~\ref{fig:motiv}. (b). This simple reformulation makes the model optimization objective become maximizing the likelihood margin. So what kind of data is more suitable for maximizing the likelihood margin? Previous methods~\cite{vlm-rlaif,ahn2024isrt,zhang2024hound} select pair data from multiple responses of the model itself, but the difference in the likelihood of such pair data is not obvious, which may make the optimization inefficient. To this end, we propose the Lean Preference Optimization (\methodname{}) framework, and a key component is a reward-trustworthiness correlated self-generated preference data pipeline, which injects relevant prior knowledge into the model while continuously refining the training data via self-reflection. This allows the policy model to estimate response likelihoods precisely. By ensuring that the collected responses maintain a consistent level of reward (likelihood) and trustworthiness, our approach fosters more reliable and well-calibrated preference learning.

Specifically, given public video-text training data (e.g., Video-QA, Video-Caption), an intuitive idea is to feed the ground-truth answers to the Video-LLMs as hints, giving the model more trustworthiness in its responses. We feed the ground-truth answer as the hint to the model along with the video to prompt the model to make a response first and then improve the response through self-reflection, to obtain a preferred (winning) response. Further, to expose the differences in trustworthiness between responses, when generating the inferior (losing) responses, we apply video augmentation to the original videos, making the responses have some randomness and fuzziness. 
Besides, since the video contents are always diverse and contain more task-independent information, a dynamic label smoothing method is introduced to mitigate the impact of potential noise. As shown at the bottom of Fig.~\ref{fig:motiv}. (b), \methodname{} makes reward trend in preference learning more stable, and the reward margin between winning and losing responses exists at the beginning and continues to expand, thus prompting the efficient alignment of Video-LLMs.
Overall, our contributions are as follows:
\begin{itemize}
    \item We revisit the likelihood displacement phenomenon in Video-LLMs, and propose a simple yet effective \methodname{} that makes the rewards learning more stable and expands the margin between winning and losing responses;
    \item We reformulate the reward of Video-LLMs as the average likelihood to mitigate likelihood displacement, and propose a reward-trustworthiness-correlated preference data pipeline that synergizes well with this new formulation to achieve efficient alignment;
    \item Extensive experiments demonstrate the effectiveness of \methodname{} on six challenging video benchmarks. Specifically, implementing our method with LLaVA-NeXT-Video-7B yields a significant improvement of 10.8\% on NeXT-QA and 12.4\% on Video-MME. 
\end{itemize}

\begin{figure*}
    \centering
    \includegraphics[width=\textwidth]{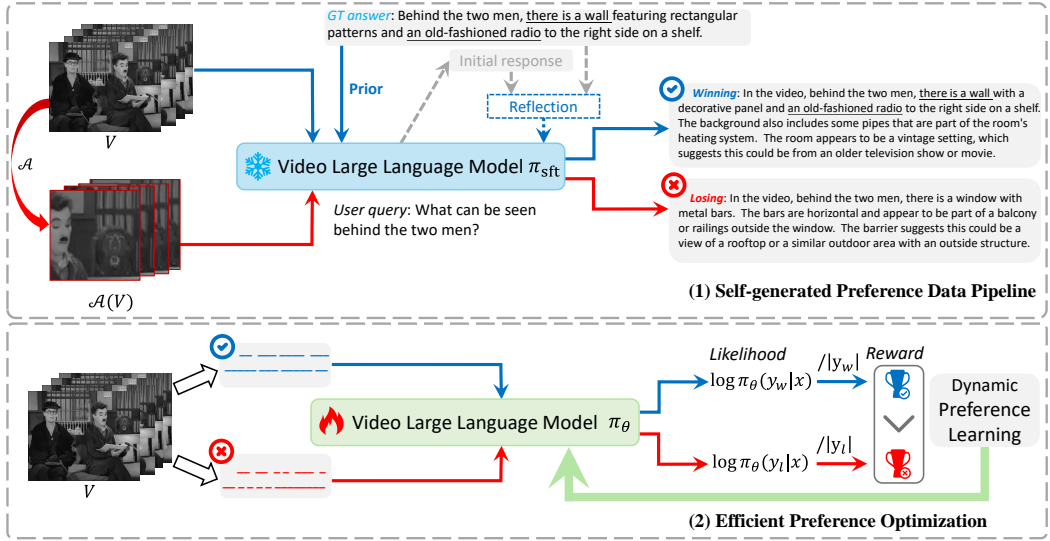}
    \caption{Overview of the \methodname{} (Lean Preference Optimization) framework. (1) Given an input video, the system prompt, and the user query, a novel method that injects the prior (the ground-truth answer with high trustworthiness) into the Video-LLM to generate the winning response, which captures the trustworthiness from the standard answer, and an losing response is generated by augmented video without prior injection. (2) Paired responses are assigned rewards by likelihoods, and we optimize the model by maximizing the reward discrepancy with dynamic preference labels. }
    \label{fig:method}
    \vspace{-5mm}
\end{figure*}


\section{Related Work}

\subsection{Likelihood Displacement Phenomenon}
Preference alignment methods like DPO~\cite{rafailov2024dpo} optimize reward margins between chosen ($y_w$) and rejected ($y_l$) responses. However, studies show both $\log\pi_\theta(y_w|x)$ and $\log\pi_\theta(y_l|x)$ often decrease during training, which a phenomenon termed \textit{likelihood displacement}~\cite{razin2024unintentional}. This contradicts the expected divergence between preferred and dispreferred outputs, while unintentionally increasing probabilities for non-target responses. Theoretical analyses attribute this behavior to three primary factors: (1) \textit{Model capacity bottlenecks} that restrict simultaneous optimization of preference margins and likelihood preservation~\cite{tajwar2024preference}; (2) \textit{Multi-token interference} from conflicting gradient signals across diverse training samples~\cite{pal2024smaug}; and (3) \textit{Suboptimal initialization} inherited from supervised fine-tuning (SFT) phases~\cite{rafailov2024r}. Critically, likelihood displacement not only diminishes the absolute probabilities of target responses but also inadvertently elevates the likelihood of \textit{non-target responses}, which are unrelated to the preference pairs—thereby degrading generalization performance~\cite{yang2025dpo,liu2024provably}. While recent work in LLMs proposes dataset filtering~\cite{razin2024unintentional} or regularization techniques~\cite{liu2024provably} to mitigate this issue, these solutions compromise either training efficiency or the simplicity that defines DPO-based methods. We argue that video data with temporal redundancies amplify likelihood displacement through preference alignment strategy in Video-LLMs. This unaddressed challenge likely hinders preference alignment efficiency in Video-LLMs, demanding an urgent investigation.

\subsection{Preference Optimization for LLMs} Large Language Models (LLMs) are typically pre-trained in an unsupervised manner on extensive textual corpora, followed by supervised fine-tuning (SFT) on high-quality instruction datasets to acquire task-specific capabilities~\cite{chatgpt,gpt4,touvron2023llama2,qwen2}. While SFT enhances alignment with human preferences, it suffers from high computational costs and potential propagation of undesirable behaviors from the training data. Reinforcement learning (RL)-based preference optimization methods, such as RLHF~\cite{christiano2017rlhf} and RLAIF~\cite{bai2022rlaif}, address these limitations by aligning models with human preferences through feedback signals, significantly improving conversational and coding abilities. Despite their effectiveness, RL-based approaches require complex pipelines involving multiple LLMs and substantial computational resources. DPO~\cite{rafailov2024dpo} pioneers a reward-model-free framework by reformulating RLHF objectives into a binary cross-entropy loss. Subsequent advancements have introduced more streamlined preference optimization approaches: PRO~\cite{song2024pro} and LiPO~\cite{liu2024lipo} leverage ranked response lists rather than pairwise comparisons, while ORPO~\cite{hong2024orpo} and SimPO~\cite{meng2024simpo} eliminate reference model dependencies, aligning optimization with generation dynamics. KTO~\cite{ethayarajh2024kto} operates without pairwise preference data by leveraging absolute quality judgments. $\beta$-DPO~\cite{wu2024betadpo} introduces a dynamic $\beta$ for all likelihoods into DPO. In this paper, we treat the likelihood of winning and losing differently and use label-smoothing to achieve dynamic preference alignment. In general, the role of different algorithms or data in Video-LLMs has not been fully explored.


\subsection{Preference Optimization for Multimodal LLMs} Multimodal LLMs (MLLMs) face exacerbated alignment challenges, as misalignment often manifests as visual hallucinations, responses ungrounded in input images or videos. Recent work adapts LLM preference optimization techniques to MLLMs through direct architectural inheritance, exemplified by RLHF-V~\cite{yu2024rlhfv} and RLAIF-V~\cite{yu2024rlaifv}. Current approaches for image-based MLLMs focus on two strategies: (1) Enhanced preference data curation through diverse response collection (Silkie~\cite{li2023silkie}), hallucination-aware annotations (RLHF-V~\cite{yu2024rlhfv}), or hybrid scoring (CSR~\cite{zhou2024csr}), and (2) Adversarial training via image distortion (POVID~\cite{zhou2024povid}, BPO~\cite{pi2025bpo}) or augmentation-based negative sampling (SeVa~\cite{zhu2024seva}, STIC~\cite{deng2024stic}, V-DPO~\cite{xie2024vdpo}). While effective for hallucination reduction, these methods show limited impact on general capability enhancement~\cite{xiong2024llavacritic}. Video-based MLLMs present unique challenges due to temporal dynamics and computational constraints. Initial efforts like VLM-RLAIF~\cite{vlm-rlaif} employ Proximal Policy Optimization (PPO) with self-evaluated rewards, while LLaVA-Hound~\cite{zhang2024hound} leverages GPT-4V~\cite{gpt4v} for automated sentence-level scoring. Current video alignment methods~\cite{ahn2024isrt,li2025temporalPO} predominantly adopt DPO, but suffer from misalignment between training objectives (pairwise preference ranking) and inference metrics (absolute generation quality). Our work addresses this gap by reformulating rewards of Video-LLMs as normalized likelihood of policy models, synergizing with self-generated preference data for efficient temporal alignment.

\section{Lean Preference Optimization}

\begin{wrapfigure}{r}{0.5\linewidth}
\vspace{-5mm}
    \centering
    \includegraphics[width=\linewidth]{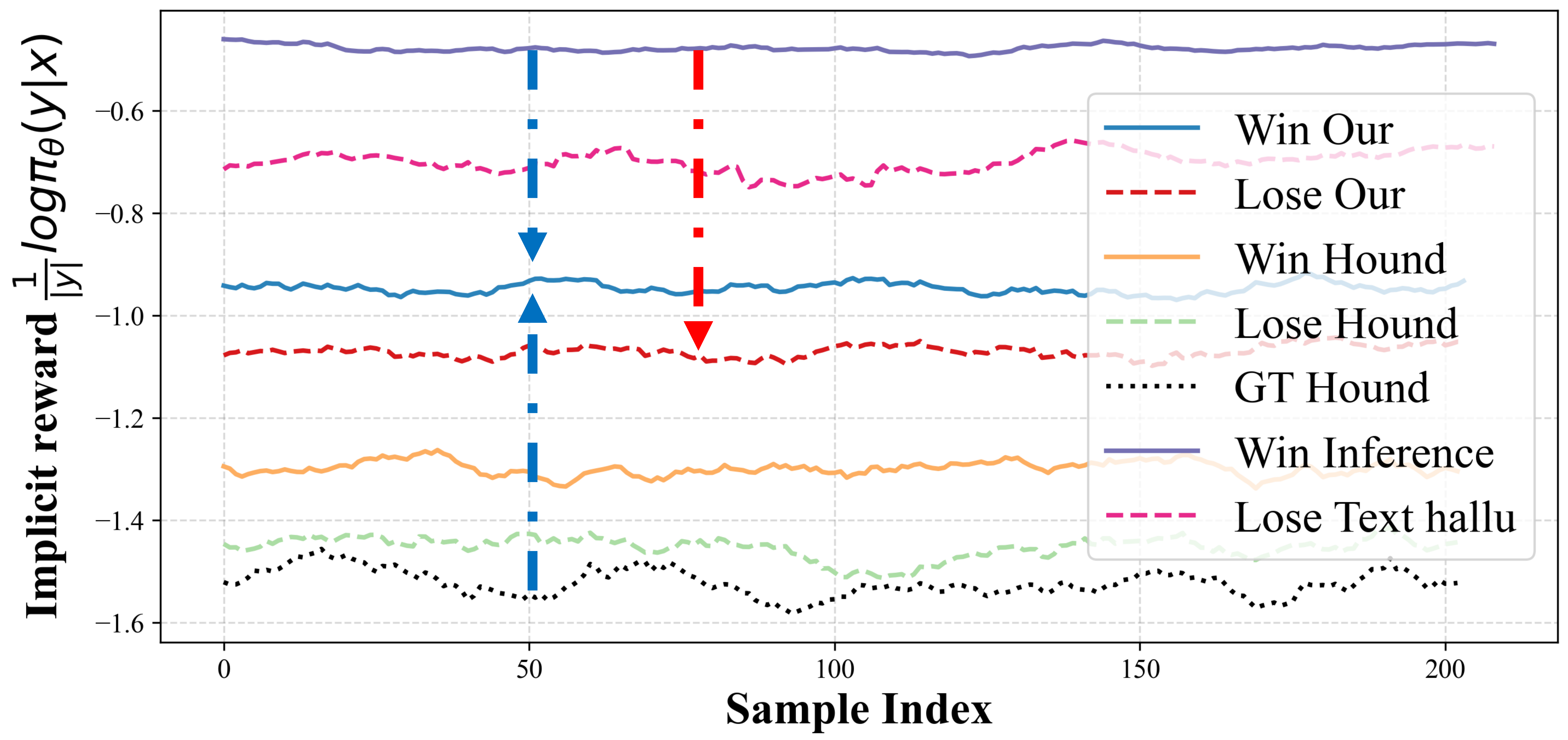}
    \vspace{-5mm}
    \caption{Implicit rewards for different responses.}
    \label{fig:find}
\vspace{-10pt}
\end{wrapfigure}

\paragraph{Motivation.} 
In Fig.~\ref{fig:motiv}. (b), the upper part illustrates the likelihood displacement phenomenon in Video-LLM alignment, where the rewards for both winning and losing responses decline, unintentionally elevating the rewards of non-target responses. As the policy model's likelihood for given responses continues to drop, the gap with the SFT model's likelihood widens. To mitigate this issue, we reformulate the reward as the average likelihood of the response under the policy model. The data distribution determines the likelihood distribution. To explore the likelihood of different responses, we analyze LLaVA-NeXT-Video~\cite{zhang2024llavanextvideo} using the LLaVA-Hound~\cite{zhang2024hound} dataset. As shown in Fig.~\ref{fig:find}, we plot reward (average likelihood) curves for ground-truth (GT), direct inference, hallucination (generated with misleading text~\cite{deng2024stic}), and Hound's original winning and losing responses. Despite high trustworthiness of GT responses, they receive the lowest rewards due to low model likelihood. Since preference training aims to enlarge the reward margin between winning and losing, we propose selecting training pairs by considering both reward and trustworthiness. Concretely, we guide the model to generate a high-reward, GT-informed response (blue arrow), and use video augmentation to produce a slightly weaker one (red arrow), and details of the pipeline are in Fig.~\ref{fig:method}.



\subsection{Reward-Trustworthiness Correlated Self-generated Preference Data Pipeline}
\label{subsec:response}
Preference learning needs feedback collected in the form of comparison pairs, where each pair includes a winning response $y_w$ and a losing response $y_l$ to the same input including the video $V$ and prompt $q$. The widely used method to collect paired data in LLMs is randomly sampling several candidates from LLMs with a large temperature, a high top-p parameter, or different random seeds. Recently, some work~\cite{yu2024rlaifv,zhang2024hound} on multi-modal LLMs' preference optimization also utilizes the same strategy. Random sampling can collect numerous responses from the model itself, but a key problem lies in how to label the winning and losing responses from candidates. So, some work use human feedbacks~\cite{yu2024rlhfv}, GPT judgment~\cite{zhang2024hound}, open-source proprietary models~\cite{yu2024rlaifv}, or attention values inside models~\cite{liu2024miadpo}. However, these methods are expensive or inefficient. In the meantime, in the video understanding scenario, only random sampling makes it hard to expose the genuine difference between responses. Also, these methods only consider the trustworthiness, so the selected winning responses may have low likelihood, resulting in poor training efficiency.
\begin{wraptable}{r}{0.58\textwidth}
\vspace{-15pt}
\caption{Prompt for winning response generation.}
\vspace{-1mm}
\centering
\small
\begin{tabular}{|p{0.95\linewidth}|}
\hline
\rowcolor[gray]{0.8}
\textbf{Prompt for Winning Response Generation} \\ \hline
\rowcolor[gray]{0.95}
Question: A chat between a curious user and an artificial intelligence assistant. The assistant gives helpful, ... \\
\rowcolor[gray]{0.95}
USER: Here are some hints: \{GT answer: $a$\} \\
\rowcolor[gray]{0.95}
Please respond based on the given hints and video content. \\
\rowcolor[gray]{0.95}
\{VIDEO\_TOKEN\} \\
\rowcolor[gray]{0.95}
\{query: $q$\} \\
\rowcolor[gray]{0.95}
ASSISTANT:\\
\rowcolor[gray]{0.95}
Response: \{Initial Response: $y_{\texttt{init}}$\}\\
\hline
\rowcolor[gray]{0.95}
// Video-LLM Reflection

USER: Your previous reply to me was:
\{Initial Response: $y_{\texttt{init}}$\}. This response can continue to be improved.

Now, please align your response with the information below:
\{GT answer: $a$\}

You need to reflect the given information as best you can, optimize your response, and enrich your answer. I'll ask you the question again:\\
\rowcolor[gray]{0.95}
\{VIDEO\_TOKEN\} \\
\rowcolor[gray]{0.95}
\{query: $q$\} \\
\rowcolor[gray]{0.95}
ASSISTANT:\\
\rowcolor[gray]{0.95}
Response: \{winning response: $y_w$\}\\
\hline

\end{tabular}
\vspace{-5mm}
\label{tab:prompt}
\end{wraptable}

To address these concerns, we propose a reward-trustworthiness-correlated principle to generate preference data, whereby the responses should have the same level of reward (likelihood) and trustworthiness. We leverage prior knowledge and a reflection method to enhance the accuracy of winning responses' information, and also try to increase the difference between paired responses. Since there is a large amount of video-text paired data in previous work, such as Video-LLM pre-training and fine-tuning~\cite{liu2024llavanext,li2024llavaonevision,zhang2024hound}, we can directly use the annotation data, including video caption or video QA data as raw material to construct preference data. A video-text instance can be denoted as $(V, q, a)$, where $V$ denotes a video, $q$ denotes a query, and $a$ denotes the ground-truth answer. 

We then introduce our data generation pipeline. To pursue a winning response with high likelihood and trustworthiness, we believe that important factors include the accuracy of the response and whether it follows the output distribution of the model. Thus, we propose a novel prior injection method, that is, we concatenate the answer $a$ with query $q$ into a simple formatted prompt as shown in the upper part of Tab.~\ref{tab:prompt}, which is fed into the Video-LLM with the video $V$:
\begin{equation}
    y_{\texttt{init}} = \pi_\texttt{sft}(V,\ [a,q]),
\end{equation}
This simple operation prompts the Video-LLM to have prior knowledge to respond to the question. Then, to avoid Video-LLM misunderstanding the information of answers or directly copy-pasting them, we encourage the model to reflect and correct its initial response $y_{\texttt{init}}$, which better ensures the rationality and information content of final output $y_w$. The detailed prompts are shown in Table~\ref{tab:prompt}.

After the prior injection and reflection, the reward of a winning response $y_w$ will be raised effectively, where the reward is defined as the average log likelihood as follows:
\begin{equation}
    r(V,q,y_w) =\frac{\beta}{|y_w|}\sum^{|y_w|}_{i=1}\log\ \pi_\texttt{sft}(y_{w,i}\ |\ [V,q], y_{w,<i}),
    \label{eq:1}
\end{equation}
where $\pi_\texttt{sft}$ is the SFT model, and $\beta$ is a constant to control the scaling of the reward.

As for generating losing responses $y_l$ with low level reward and trustworthiness. We apply some augmentations to the original videos $V$ and prompt the Video-LLM to respond to the same query. Due to the random and noise augmentation, the losing response $y_l$ will have lower trustworthiness, and it will slightly deviate from the output distribution produced by the model when it sees normal videos. The reward of losing $y_l$ is defined as follows:
\begin{equation}
    r(V,q,y_l) =\frac{\beta}{|y_l|}\sum^{|y_l|}_{i=1}\log\ \pi_\texttt{sft}(y_{l,i}\ |\ [V,q], y_{l,<i}).
    \label{eq:2}
\end{equation}

In this way, our pipeline using existing video-text data, can prompt any Video-LLM itself to generate paired preference data, and there is an obvious difference between winning and losing responses, including reward and trustworthiness. For almost any triplet $(V, y_w, y_l)$, it satisfies the $r(V,q,y_w)>r(V,q,y_l)$. The reward curve of our generated paired data can be seen in Fig.~\ref{fig:find}.

\subsection{Dynamic Preference Learning}
The data pairs generated by our data generation framework have significant differences, including in reward and trustworthiness. Thus, maximizing the reward discrepancy is easier than maximizing the difference between comparable rewards, thus benefiting the stability of optimization.
Additionally, we add a reward margin term $\gamma$, where $\gamma \geq 0$ in our training. The Bradley-Terry objective is defined as follows:
\begin{equation}
    p(y_w \succ y_l \mid [V, q]) = \sigma \left( r(V,q, y_w) - r(V,q, y_l) - \gamma \right)
    \label{eq:3}
\end{equation}

Due to the diversity of videos and the influence of some noise data, winning responses may sometimes be of worse quality than losing responses. The intuitive result is that the likelihood of the winning response is lower than the likelihood of the losing response, which may make it difficult to directly optimize the objective of Eq.~\ref{eq:3}. To alleviate the problem, we introduce a dynamic label smoothing method to correct the potential noise, where the pseudo label $z_q$ is:
\begin{equation}
z_q =
\begin{cases} 
1, & \text{if } \Big(r(V,q, y_w) - r(V,q, y_l)\Big)>d, \\
0, & \text{else}.
\end{cases}
\end{equation}
where $d$ is a hyper-parameter. Using pseudo labels, we define the dynamic preference learning objective as below:
\begin{equation}
    \Tilde{p}(y_w \succ y_l) = (1-z_q\cdot \alpha) \cdot p(y_w \succ y_l) + z_q\cdot \alpha \cdot p(y_l \succ y_w),
\end{equation}
where $p(y_w \succ y_l \mid [V, q])$ is abbreviated to $p(y_w \succ y_l)$, and $\alpha$ denotes the label smoothing factor.

Finally, using the self-generated data $\mathcal{D}$, we define the final learning objective of our framework:
\begin{equation}
    \mathcal{L}_{\text{\methodname{}}}(\pi_\theta) = - \mathbb{E}_{(V,q, y_w, y_l) \sim \mathcal{D}} \big[  \Tilde{p}(y_w \succ y_l) \big],
\end{equation}

By optimizing the \methodname{} loss on self-collected preference dataset, the model extracts richer information from input videos to enhance the trustworthiness, thus better aligning with winning responses. Consequently, Video-LLM can achieve alignment optimization more efficiently.



\section{Experiment}

\subsection{Setup}
\paragraph{Implementation Details} To verify the effectiveness of our method, we utilize two video-LLMs, including LLaVA-NeXT-Video-7B~\cite{zhang2024llavanextvideo} and LLava-Video-7B~\cite{zhang2024llavavideo}, which have different LLM backbones. We freeze the visual encoder and multi-modal projection and fine-tune all the parameters of LLM backbones. All experiments are completed on A100 GPUs, and we utilize 16 frames for full-training Video-LLMs due to the limited GPU memory. More details are shown in Appendix~\ref {app:data}.

\paragraph{Data \& Benchmarks} The annotation data are sampled from the video QA or caption data from~\cite{zhang2024hound}, including 17k data. For each SFT model, we utilize all 17k triplets (video, query, answer) to generate paired responses. For the study on preference data size, we also randomly sample annotation data from LLaVA-Video-178k~\cite{zhang2024llavavideo}. For all-round evaluation, we consider 4 multi-choice QA benchmarks: Video-MME~\cite{fu2024videomme}, LongVideoBench (LongVB)~\cite{wu2024longvideobench}, MLVU~\cite{MLVU}, NeXT-QA~\cite{xiao2021next}, and 2 open-ended QA benchmarks: DREAM-1K~\cite{wang2024tarsier} for detailed video description and VideoChatGPT~\cite{maaz2023video} for chat.







\begin{table*}[t]
\caption{Results on LongVB~\cite{wu2024longvideobench}, MLVU~\cite{MLVU}, NeXT-QA~\cite{xiao2021next}, and Video-MME~\cite{fu2024videomme} compared with state-of-the-art models. The Video-MME results are in the format ``w.o subs".}
\centering
\resizebox{\textwidth}{!}{
\begin{tabular}{lcccccccc}
\toprule
\multirow{2}{*}{\textbf{Model}} & \multirow{2}{*}{\textbf{Size}} & \multirow{2}{*}{\textbf{LongVB}} & \multirow{2}{*}{\textbf{MLVU }} & \multirow{2}{*}{\textbf{NeXT-QA }} & \multicolumn{4}{c}{\textbf{Video-MME}} \\
\cmidrule(lr){6-9}
 &  & \textbf{(val)}  &  \textbf{(M-avg)}& \textbf{(mc)} & \textbf{Short} & \textbf{Medium} & \textbf{Long} & \textbf{Overall} \\
\midrule
Video-LLaVA \cite{lin2023video} & 7B & 39.1 & 47.3 & - & 45.3 & 38.0 & 36.2 & 39.9 \\
Qwen-VL-Max \cite{Qwen-VL}&-&-&42.2  & - &55.8 &49.2 &48.9 &51.3 \\
ShareGPT4Video \cite{chen2024sharegpt4video} & 8B & 39.7 & 46.4 & -  & 48.3  & 36.3  & 35.0  & 39.9  \\
InternVL-Chat-V1.5 \cite{chen2024internvl} & 20B & 51.2 & 50.4 & - & 50.7  & \textbf{60.2}  & 46.4  & 45.6  \\
VideoChat2 \cite{2023videochat} & 7B & 39.3 & 47.9 &- & 48.3  & 37.0  & 33.2  & 39.5  \\
LongLLaVA  \cite{long-llava-qwen2-7b-2024}& 7B & - & 56.3 & - & 61.9  & 51.4  & 45.4  & 52.9  \\
Video-CCAM \cite{fei2024video}&14B&-&\underline{63.1}& - &62.2 &50.6 &46.7 &53.2 \\

LongVA~\cite{zhang2024long} & 7B & 51.3 & 58.8 & 68.3 & 61.1  & 50.4  & 46.2  & 52.6  \\


LLaVA-Video-TPO~\cite{li2025temporalPO} & 7B & \textbf{59.0} & 62.9 & 77.6 & \underline{71.3}  & 56.9  & 49.0  & 59.1  \\

\midrule
LLaVA-NeXT-Video~\cite{zhang2024llavanextvideo} & 7B & 40.1 & 43.4 & 53.9 & 44.0 
 & 38.0  & 34.4  & 38.8  \\
\rowcolor{cyan!10} \textbf{LLaVA-NeXT-Video-\methodname{}}~\footnotesize{(\textbf{\texttt{ours}})}& 7B & 44.0 & 45.1 & 59.7 & 51.4  & 43.0  & 36.6  & 43.6  \\
\rowcolor{cyan!10} $\Delta$\  (\methodname{})   & - & \textcolor{red}{10.0\%↑} & \textcolor{red}{3.9\%↑} & \textcolor{red}{10.8\%↑} & \textcolor{red}{16.8\%↑}  & \textcolor{red}{13.2\%↑}  & \textcolor{red}{6.4\%↑ } & \textcolor{red}{12.4\%↑}  \\

\midrule
LLaVA-Video~\cite{zhang2024llavavideo} & 7B & 58.5 & 62.6 & \underline{79.4} & \underline{71.3}  & 57.4  & \underline{49.6}  & \underline{59.4}  \\

\rowcolor{cyan!10} \textbf{LLaVA-Video-\methodname{}}~\footnotesize{(\textbf{\texttt{ours}})} & 7B & \underline{58.9} & \textbf{63.3} &\textbf{81.2} & \textbf{71.6}  & \underline{57.9}  & \textbf{50.2}  & \textbf{59.9}  \\
\rowcolor{cyan!10} $\Delta$\  (\methodname{}) & - & \textcolor{red}{0.7\%↑}  & \textcolor{red}{1.1\%↑} & \textcolor{red}{2.3\%↑} & \textcolor{red}{0.4\%↑}   & \textcolor{red}{0.9\%↑}  & \textcolor{red}{1.2\%↑}  & \textcolor{red}{0.8\%↑} \\
\bottomrule
\end{tabular}
}

\vspace{-5mm}
\label{tab:results}
\end{table*}
  
\subsection{Quantitative Results}

\paragraph{Multiple-choice QA benchmarks} Tab.~\ref{tab:results} presents the comparison results with current state-of-the-art video-LLMs on 4 multiple-choice QA benchmarks. With the introduction of \methodname{}, both LLaVA-NeXT-Video and LLaVA-Video significantly outperform their corresponding baselines. It brings significant improvements of 10\% on LongVB, 3.9\% on MLVU, 10.8\% on NeXT-QA, and 12.4\% on Video-MME. Based on the state-of-the-art LLaVA-Video, our method achieves better improvements against TPO~\cite{han2022temporal}. Our model outperforms all 7B baselines on MLVU, NeXT-QA, and Video-MME, and achieves a closer score to GPT-4o on MLVU. The results consistently indicate our \methodname{} potential to improve the capability on multiple-choice QA.

\paragraph{Open-ended QA benchmarks} Tab.~\ref{tab:dream} presents the comparison of detailed video description on DREAM-1K. Our method brings improvements of +3.2 and +1.2 on the F1 score for two baseline models, respectively, and our model outperforms all 7B baselines. Tab.~\ref{tab:vcg} presents the results compared with current state-of-the-art models on VideoChatGPT. CI, DO, CU, TU, and CO refer to the correctness of information, detail orientation, contextual understanding, temporal understanding, and consistency, respectively. Our method surpasses DPO on 3 out of 5 tasks, achieves a higher average performance, and significantly outperforms GPT-4V on the consistency (CO) task.

The consistent improvements over two baseline models across various multiple-choice and open-ended QA benchmarks validate the effectiveness of our method and highlight its ability to enhance the comprehensive video understanding capabilities of pretrained Video-LLMs.

\begin{table}[t]
\centering
\begin{minipage}[b]{0.47\textwidth}
    \centering
\caption{\textbf{GPT-based evaluation for video detailed captioning.} Results on DREAM-1K.}
\resizebox{\textwidth}{!}{

\centering
\resizebox{\linewidth}{!}{
\begin{tabular}{lccc}
\toprule[1.pt]
\textbf{Model} & \textbf{F1} & \textbf{Precision} & \textbf{Recall} \\
\midrule
Video-LLaVA \cite{lin2023video} & 20.4 & 28.1 & 16.0 \\
VideoChat2~\cite{li2023mvbench} & 26.6 & 31.0 & 23.3 \\
MiniGPT-4Video~\cite{ataallah2024minigpt4video} & 24.0 & 26.1 & 22.2 \\
PLLaVA-34B~\cite{xu2024pllava} & 28.2 & \textbf{38.4} & 22.3 \\
LLaVA-OV-7B~\cite{li2024llavaonevision} & 31.7 &  -  & - \\
\textcolor{gray}{LLaVA-OV-72B}~\cite{li2024llavaonevision} & \textcolor{gray}{33.2} &  \textcolor{gray}{-}  & \textcolor{gray}{-} \\

\midrule
LLaVA-NeXT-Video~\cite{zhang2024llavanextvideo} & 23.3 &31.1 &18.6 \\
\rowcolor{cyan!10} \textbf{LLaVA-NeXT-Video-\methodname{}}~\footnotesize{(\textbf{\texttt{ours}})} & 26.5  & 34.6   & 21.5  \\

\midrule
LLaVA-Video~\cite{zhang2024llavavideo} & 31.7 & 33.6 & 29.9 \\
\rowcolor{cyan!10} \textbf{LLaVA-Video-\methodname{}}~\footnotesize{(\textbf{\texttt{ours}})} & \textbf{32.9}  & 33.5   & \textbf{32.4}  \\

\bottomrule[1.pt]
\end{tabular}
}
\label{tab:dream}
}
\end{minipage}
\hspace{0.01\textwidth}
\begin{minipage}[b]{0.5\textwidth}
    \centering
    \caption{\textbf{GPT-based evaluation for video chat.} Results on VideoChatGPT~\cite{maaz2023video}.}
\label{tab:vcg}
\resizebox{\textwidth}{!}{
\centering
\resizebox{0.7\linewidth}{!}{
\begin{tabular}{lcccccc}
\toprule[1.pt]
\textbf{Methods} & \textbf{CI} & \textbf{DO} & \textbf{CU} & \textbf{TU} & \textbf{CO} & \textbf{Avg.} \\
\midrule
 VideoChat~\cite{li2023videochat} & 2.23 & 2.50 & 2.53 & 1.94& 2.24 &2.29 \\
 Video-ChatGPT~\cite{maaz2023video} & 2.50 & 2.57 & 2.69 &2.16 &2.20 &2.42 \\ 
 Vista-LLaMA\cite{ma2024vista} & 2.44 & 2.64 & 3.18 & 2.26 & 2.31 & 2.57 \\
 MoiveChat~\cite{song2024moviechat} & 2.76 & 2.93 & 3.01 & 2.24 & 2.42 & 2.67 \\
Chat-UniVi~\cite{jin2023chat-univi} & 2.89 & 2.91 & 3.46 & 2.40 & 2.81 & 2.89  \\
VideoChat2~\cite{li2023mvbench} & 3.02 & 2.88 & 3.51 & 2.66 & 2.81 & 2.98 \\

PLLaVA~\cite{xu2024pllava} & 3.21 & 2.86 & 3.62 & 2.33 & 2.93 & 2.99 \\

CAT~\cite{ye2024cat} & 3.08 &2.95 &3.49 &2.81 &2.89 &3.07 \\
ST-LLM~\cite{liu2024st} & 3.23 & 3.05 &3.74 & 2.93 & 2.81 &3.15 \\


\midrule
LLaVA-NeXT-Video~\cite{zhang2024llavanextvideo} & 3.39 &3.29 &3.92 &2.60 &3.12 &3.26 \\
LLaVA-NeXT-Video-DPO~\cite{rafailov2024dpo} & 3.64 &3.45 & \textbf{4.17} & \textbf{2.95} & 4.08 &3.66 \\
\rowcolor{cyan!10} \textbf{LLaVA-NeXT-Video-\methodname{}}~\footnotesize{(\textbf{\texttt{ours}})} & \textbf{3.77}  & \textbf{3.51}   & 4.14  & 2.72  &  \textbf{4.26}  & \textbf{3.68} \\

\bottomrule[1.pt]
\end{tabular}
}
}
\end{minipage}
\end{table}

\subsection{Ablation Study}

\paragraph{Comparison of different algorithms and different data} In Tab.~\ref{tab:dpo}, we compare our method with SFT, DPO, SimPO on various settings, using 17k data from~\cite{zhang2024hound}. When comparing methods using GT data, DPO and SimPO under the GT-as-Winning setting achieve lower scores on Video-MME than SFT. In contrast, our framework outperforms DPO and SimPO, achieving the highest scores on Video-MME, LongVB, MLVU, and DREAM-1K. Additionally, by leveraging our data generation pipeline, both DPO and SimPO have significant improvements on these benchmarks, further validating the effectiveness of our data generation pipeline. In Tab.~\ref{tab:data2}, we test \methodname{} using different responses. Compared with responses of GT or other models~\cite{zhang2024hound}, or use hallucination response (using misleading text)~\cite{deng2024stic}, \methodname{} achieves the best performance using data from our data generation pipeline.



\begin{figure}[]
    \centering
    \begin{subfigure}[b]{0.49\linewidth} 
        \includegraphics[width=\linewidth, height=6.5cm]{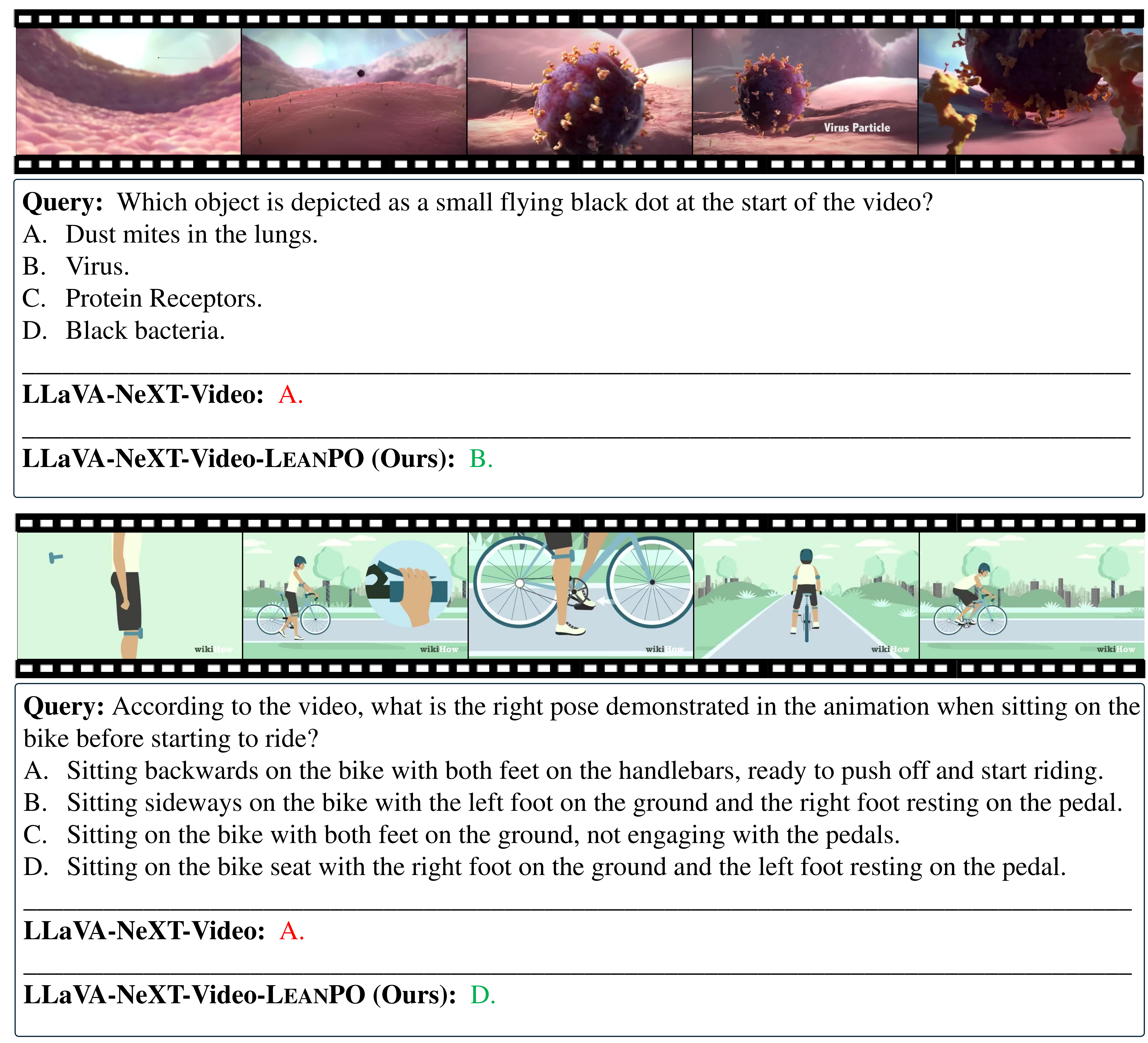} 
        \caption{Video QA: The correct answer is marked in green.}
        \label{fig:dream1a}
    \end{subfigure}
    \hfill 
    \begin{subfigure}[b]{0.49\linewidth} 
        \includegraphics[width=\linewidth]{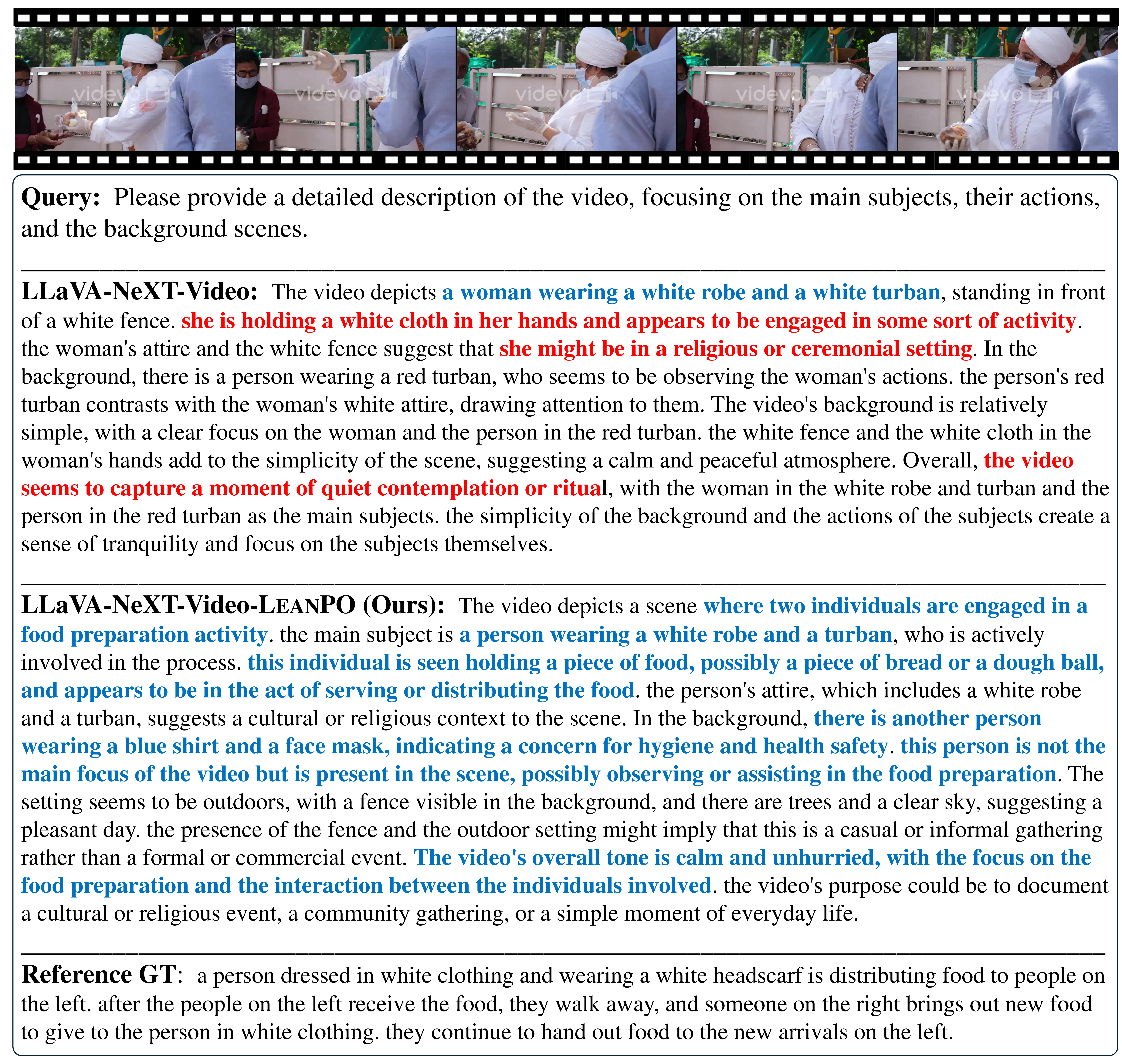} 
        \caption{Video Detail Captioning}
        \label{fig:dream1b}
    \end{subfigure}
    \caption{Qualitative comparison for Video Detail Captioning and Video QA. \textcolor{red}{Red} indicates irrelevant responses, while \textcolor{blue}{blue} indicates well-grounded responses.}
    \label{fig:dream1}
\vspace{-10pt}
\end{figure}
\begin{table*}
\caption{Comparison between other preference learning methods and our proposed \methodname{}.}
\centering
\resizebox{\textwidth}{!}{
\begin{tabular}{l|ccc|ccccccc}
\toprule[1.pt]
 \multirow{2}{*}{\textbf{Method}} & \multirow{2}{*}{\textbf{Data source}} &  \multirow{2}{*}{\textbf{Win}} &  \multirow{2}{*}{\textbf{Lose}}   & \multicolumn{4}{c}{\textbf{Video-MME}} & \multirow{2}{*}{\textbf{LongVB}} & \multirow{2}{*}{\textbf{MLVU }} & \multirow{2}{*}{\textbf{DREAM-1K }} \\
\cmidrule(lr){5-8}
& & &  & \textbf{Short} & \textbf{Medium} & \textbf{Long} & \textbf{Overall}  & \textbf{(val)}& \textbf{(M-avg)} & \textbf{(test)}  \\

\midrule

Base & - & - & - & 44.0 & 38.0  & 34.4  & 38.8 & 40.1  & 43.4  & 23.3  \\

- SFT & Hound & GT & - &   46.2 & 40.4 & 34.9 & 40.5  & 40.8 & 43.0 & 24.8 \\

- DPO & Hound & Win & Lose &  48.7 & 39.6  &  35.9  & 41.4  & 42.0  & 43.6 & 19.0 \\

- DPO & Hound & GT & Lose & 43.9   & 38.7   & 33.6   &  38.7  & 40.6 & 42.8  & 25.8 \\
- DPO & Our pipe & Win & Lose  & 49.4  & 42.8  & 35.0  & 42.4  & 43.0  & 44.2 &  26.3 \\
\midrule

- SimPO & Hound & Win & Lose  & 44.9  & 39.7  & 34.1  & 39.6  & 40.9  & 45.0  & 26.1 \\

- SimPO & Hound & GT & Lose & 43.7   & 38.6   &  32.0   &  38.1  & 43.0  & 43.8 & 26.0 \\

- SimPO & Our pipe & Win & Lose  & 50.6  & 42.1  & 35.0  & 42.6  & 43.5  & 40.8 &  26.1 \\
\midrule

\rowcolor{cyan!10} - \methodname{} & Our pipe & Win & Lose & \textbf{51.4}   &  \textbf{43.0}   &  \textbf{36.6}   & \textbf{43.6}  & \textbf{44.0} & \textbf{45.1} & \textbf{26.5} \\

\bottomrule[1.pt]
\end{tabular}
}

\label{tab:dpo}
\end{table*}
\begin{table}
\vspace{-5mm}
\caption{Ablation study of \methodname{} with LLaVA-NeXT-Video-7B on various benchmarks.}
\centering
\resizebox{\textwidth}{!}{
\begin{tabular}{l|ccc|ccccccc}
\toprule[1.pt]
\multirow{2}{*}{\textbf{Method}} & \multirow{2}{*}{\textbf{Self-gen data}} & \multirow{2}{*}{\textbf{Reflect}} & \multirow{2}{*}{\textbf{Dynamic}}  & \multicolumn{4}{c}{\textbf{Video-MME}} & \multirow{2}{*}{\textbf{LongVB}} & \multirow{2}{*}{\textbf{MLVU }} & \multirow{2}{*}{\textbf{DREAM-1K }} \\
\cmidrule(lr){5-8}
& & & & \textbf{Short} & \textbf{Medium} & \textbf{Long} & \textbf{Overall}  & \textbf{(val)}& \textbf{(M-avg)} & \textbf{(test)}  \\

\midrule

Base & - & - & - & 44.0  & 38.0   & 34.4   & 38.8  & 40.1  & 43.4  & 23.3  \\

DPO & - & - & - & 48.7 & 39.6  &  35.9  & 41.4  & 42.0  & 43.6 & 19.0 \\
SimPO & - & - & - & 44.9  & 39.7  & 34.1  & 39.6  & 40.9  & 45.0  & 26.1 \\
\midrule
\multirow{3}{*}{\methodname{}} & \cmark & & & 50.6  & 42.1  & 35.0  & 42.6  & 43.5 &40.8 & 26.1 \\ 
& \cmark & \cmark & & 51.0  & 42.4  & 36.3  & 43.2  & 43.9 & 44.5 & 25.3 \\ 
& \cmark & \cmark & \cmark & \textbf{51.4}   &  \textbf{43.0}   &  \textbf{36.6}   & \textbf{43.6}  & \textbf{44.0} & \textbf{45.1} & \textbf{26.5} \\

\bottomrule[1.pt]
\end{tabular}
}

\label{tab:ablatin}
\end{table}

\begin{table}[t]
\centering
\vspace{-4pt}
\begin{minipage}[b]{0.45\textwidth}
    \centering
    \caption{Label smoothing factor $\alpha$ analysis.}
    \label{tab:alpha}
\resizebox{\textwidth}{!}{
\begin{tabular}{l|cccc}
\toprule
\multirow{2}{*}{\textbf{Model}}  & \multicolumn{4}{c}{\textbf{Video-MME}}  \\
\cmidrule(lr){2-5}
 & \textbf{Short} & \textbf{Medium} & \textbf{Long} & \textbf{Overall}  \\

\midrule

Base & 44.0  & 38.0   & 34.4   & 38.8   \\

\methodname{}$_\textbf{$\alpha$=0.5}$ & 50.8  & 42.1  & 36.0  & 43.0   \\ 

\methodname{}$_\textbf{$\alpha$=0.3}$ & 50.8  & 42.7  & 36.0  & 43.2   \\ 

\rowcolor{cyan!10} \methodname{}$_\textbf{$\alpha$=0.1}$ & \textbf{51.4}   &  \textbf{43.0}   &  \textbf{36.6}   & \textbf{43.6}  \\

\bottomrule
\end{tabular}
}
\end{minipage}
\hspace{0.01\textwidth}
\begin{minipage}[b]{0.52\textwidth}
    \centering
    \caption{Different response effect analysis.}
    \label{tab:data2}
\resizebox{\textwidth}{!}{
\begin{tabular}{cc|cccc}
\toprule
\multirow{2}{*}{\textbf{Win}} & \multirow{2}{*}{\textbf{Lose}}  & \multicolumn{4}{c}{\textbf{Video-MME}}  \\
\cmidrule(lr){3-6}
& & \textbf{Short} & \textbf{Medium} & \textbf{Long} & \textbf{Overall}  \\

\midrule

Hound Win & Hound Lose  & 48.3   &  40.2  &  36.0  &  41.5  \\

Hound GT & Hound Lose &  49.8 &  39.9 &  34.3 & 41.3   \\ 

Inference & Text-hallu & 47.0  & 39.2  & 34.8  & 40.3   \\ 


\rowcolor{cyan!10} Our Win & Our Lose  & \textbf{51.4}   &  \textbf{43.0}   &  \textbf{36.6}   & \textbf{43.6}  \\

\bottomrule
\end{tabular}
}
\end{minipage}
\vspace{-4mm}
\end{table}


\paragraph{Framework components}Tab.~\ref{tab:ablatin} illustrates the impact of the data generation pipeline, reflection, and dynamic labels. Using our data, \methodname{} gets better scores than other baselines on Video-MME and LongVB. By incorporating reflection and dynamic techniques, it achieves overall improvements rather than enhancing only specific capabilities. Tab.~\ref{tab:alpha} presents the effects the the smoothing factor for dynamic labels, where the most commonly used value of 0.1 has the best effect.

\begin{wrapfigure}{r}{0.5\linewidth}
    \centering
    \includegraphics[width=\linewidth]{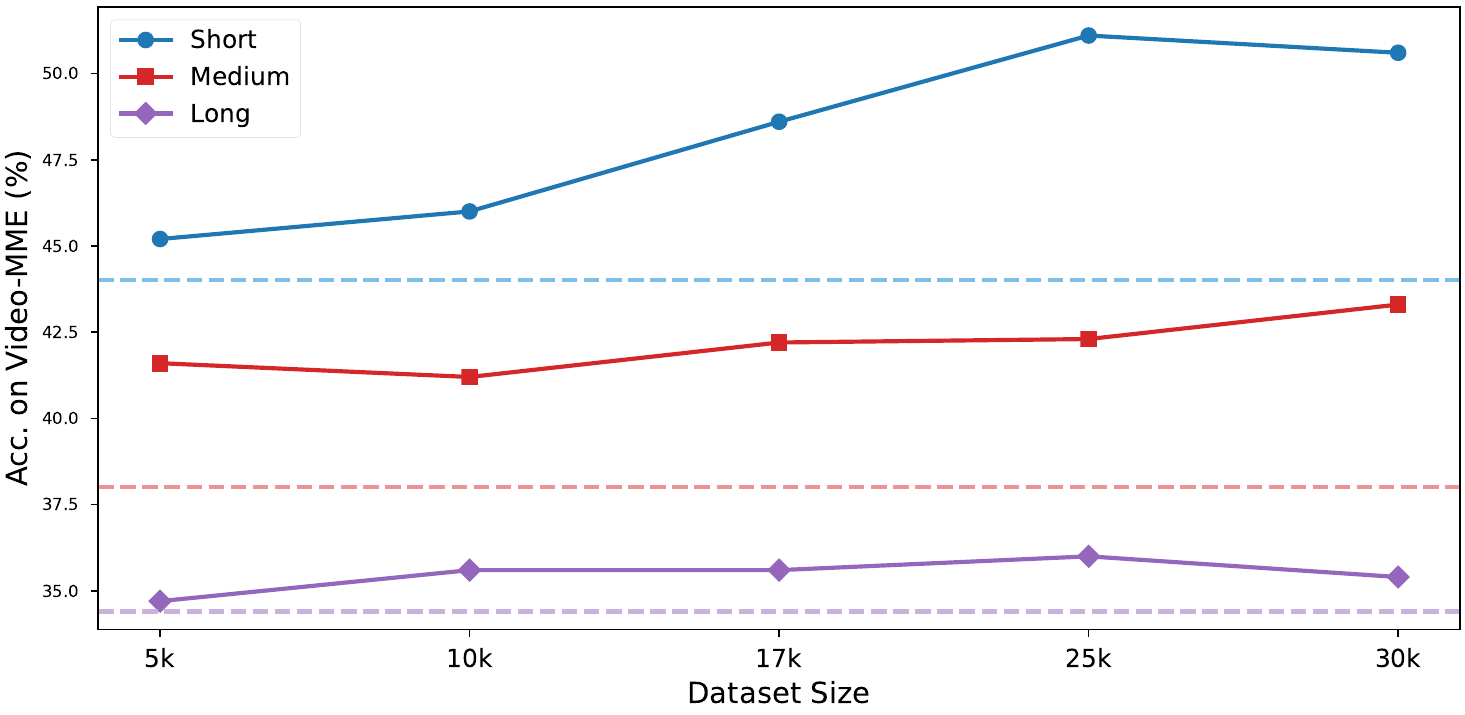}
    \vspace{-3mm}
    \caption{Impact of self-generated preference dataset size based on LLaVA-NeXT-Video-7B~\cite{zhang2024llavanextvideo} for on Video-MME. The present results are in the format ``w.o subs". The model only traverses the preference dataset once during training.}
    \label{fig:mmesize}
\vspace{-40pt}
\end{wrapfigure}

\vspace{-2pt}
\paragraph{Effect of preference dataset size}In Fig.~\ref{fig:mmesize}, we examine the scalability of our \methodname{}. To emphasize the importance of data quantity, we traverse datasets of different sizes only once during training. 
The intuitive result is that increasing the data generally improves average performance on Video-MME. However, with larger video datasets, certain specific capabilities may improve further, as evidenced by higher scores in medium-duration video understanding.



\subsection{Qualitative Results}
The qualitative comparison results between our method and baseline on Video QA and detailed video description are shown in Fig.~\ref{fig:dream1a} and Fig.~\ref{fig:dream1b}. 
In the first example, which requires models to link the flying black dot in the former frames to the virus in the latter frames, our model successfully comprehends and links the related content. In the second example, which requires models to recognize the action correctly, the baseline misidentifies continuous actions in the video. In contrast, our model accurately identifies the actions and gives accurate responses. Fig.~\ref{fig:dream1b} shows an example of detailed video description from DREAM-1K. For the original LLaVA-NeXT-Video, it only sees the appearance features of the main object, and the rest of the information means a lot of hallucinations. In contrast, our model not only sees the appearance features of the main object and other objects, but also infers a very accurate and relevant description of the video content.


\subsection{Generalization of LenPO on Video-LLMs}
\begin{table*}[]
\caption{Qwen-VL Series Video-LLMs performance using our proposed \methodname{}.}
\resizebox{\textwidth}{!}{
\begin{tabular}{l|ccc|cccccc}
\toprule[1.pt]
 \multirow{2}{*}{\textbf{Method}} & \multirow{2}{*}{\textbf{Data source}} &  \multirow{2}{*}{\textbf{Win}} &  \multirow{2}{*}{\textbf{Lose}}   & \multicolumn{4}{c}{\textbf{Video-MME}} & \multirow{2}{*}{\textbf{MLVU}} & \multirow{2}{*}{\textbf{LongVideoBench}}  \\
\cmidrule(lr){5-8}
& & &  & \textbf{Short} & \textbf{Medium} & \textbf{Long} & \textbf{Overall}  & \textbf{(val)} & \textbf{(test)}   \\

\midrule

Qwen2-VL-7B~\cite{wang2024qwen2vl} & - & - & - & 68.1 & 52.3  & 47.8 & 56.0 & 59.8  & \textbf{55.9}   \\

- SFT & Hound & GT & - & 67.5 & 54.7 & 48.9 & 57.0 & 60.0 &  53.1   \\

- DPO & Hound & Win & Lose &  70.2 & 54.9  &  48.4  & 57.8  & 59.5  & 55.2 \\

\rowcolor{cyan!10} - \methodname{} & Ours & Win & Lose & \textbf{70.8}   &  \textbf{56.4}   &  \textbf{48.8}   & \textbf{58.7}  & \textbf{60.3} & \textbf{55.9}  \\ 


\midrule
Qwen2.5-VL-7B~\cite{bai2025qwen25vl} & - & - & - & 70.6 & 56.8 & 49.2 & 58.9 & 61.5 & 59.0  \\
- SFT & Hound & GT & - & 70.4  &  57.6 & 51.1 & 59.7 & 60.0  &  57.1  \\

- DPO & Hound & Win & Lose  & 72.3  & 57.1  & 49.9  & 59.8  & 60.9  & \textbf{59.2}  \\

\rowcolor{cyan!10} - \methodname{} & Ours & Win & Lose & \textbf{73.9}   &  \textbf{59.7}   &  \textbf{51.5}   & \textbf{61.7}  & \textbf{61.7} & 59.1  \\ 

\bottomrule[1.pt]
\end{tabular}
}

\label{tab:qwen}
\end{table*}

\paragraph{Details}
We conduct experiments on the latest Qwen-VL series Video-LLMs to validate the generalization ability of our proposed \methodname{}. The training video data originates from LLaVA-Hound~\cite{zhang2024hound}. In Table.~\ref{tab:qwen}, we present the results of Qwen2-VL-7B~\cite{wang2024qwen2vl} and Qwen2.5-VL-7B~\cite{bai2025qwen25vl} using different post-training methods. All training and evaluation experiments use 32 video frames. We test these models on three benchmarks, including Video-MME~\cite{fu2024videomme}, MLVU~\cite{MLVU}, LongVideoBench~\cite{wu2024longvideobench}. Video-MME has three types of duration (short, medium, long), and focuses more on temporal reasoning tasks. MLVU and LongVideoBench focus more on fine-grained details and long video understanding.

Compared with SFT and DPO~\cite{rafailov2024dpo}, our proposed \methodname{} achieves large improvements over the performance of the two baselines. In Video-MME, which focuses more on temporal reasoning tasks, \methodname{} has significant improvements. However, in MLVU and LongVideoBench, which focus more on details and long video understanding, training the above methods with limited frames has no obvious improvement. Nevertheless, \methodname{} maintains the performance better than SFT and DPO.

\begin{table*}[!t]
\caption{Generalization results of our method on Image-LLMs. We show a comparison with other preference algorithms on LLaVA-v1.5-7B~\cite{liu2023visual}. We report the results on 4 comprehensive multimodal benchmarks for reference. The best and second-best results are shown in \textbf{bold} and \underline{underlined} respectively.}
    \centering
    \resizebox{\linewidth}{!}{
    \begin{tabular}{l|c|cccc|ccccc}
    \toprule
      Method   & \#Prompts & WildVision & LLaVA-W  & LiveBench & L-Wilder &  $\text{MME}^{P}$ & $\text{MME}^{C}$ & MMB-en & MM-Vet & MMStar \\
    \midrule
      LLaVA-v1.5-7B  & -- & 24.0 & 63.4  & 39.0 & 53.0 & \underline{1510.7} & 348.2 & 64.0 & 31.1 & 33.3 \\
      
      + SIMA~\cite{wang2024SIMA} & 17k & 17.6 &  66.1  & \underline{43.9} &  52.3 & 1507.7 & \textbf{379.3} & \underline{64.9} & 31.6 & \underline{34.7}\\
      
      + CSR~\cite{zhou2024csr} & 15k & 20.0 & 71.1  & 42.6 & 55.9 & \textbf{1524.2} & \underline{367.9} & \textbf{65.4} & \underline{33.9} & 33.6\\
      
      + RLAIF-V~\cite{yu2024rlaifv} & 33.8k & 19.2 & \underline{72.7}  & \textbf{44.8} &  \underline{56.4} & 1362.7 & 302.9 & 62.6 & 26.7 & \textbf{35.4} \\
      + RLHF~\cite{sun2023llavarlhf} & 9.4k & 19.8 & 63.7  & - &  54.5 & 1508.2 & 360.2 & 60.4 & 31.1 & 33.0\\
      
      + LLaVA-Critic~\cite{xiong2024llavacritic} & 9.4k & \underline{29.2} & \textbf{73.5}  & - & \textbf{57.2} & 1500.4 & 350.7 & 64.1 & 32.2 & 34.2  \\
      
\rowcolor{cyan!10}   + \methodname{} & 9.4k & \textbf{32.1} & 71.1   & 42.6  & 54.7 & 1480.6  & 326.4  &  \underline{64.9}  & \textbf{36.7}  & 34.2  \\


    \bottomrule
    \end{tabular}
    }
    
    \label{tab:dpo_llava15}
\end{table*}

\subsection{Generalization of LeanPO on Image-LLMs}
Generalization results of our method on Image-LLMs. We show a comparison with other preference algorithms on LLaVA-v1.5-7B~\cite{liu2023visual}. We report the results on 9 comprehensive multimodal benchmarks for reference. Compared with other methods, our method can achieve comparable performance improvements with less data. In particular, we achieve the best scores on both WildVision and MM-Vet. Compared with the baseline, our method has a significant performance improvement on 7 benchmarks, which shows the generalization of our method.

\section{Conclusion}
In this paper, we present Lean Optimization Alignment (\methodname{}) to creatively address the limitation of preference alignment in Video-LLMs. We first revisit the likelihood displacement in Video-LLMs alignment. To alleviate the adverse effects, we reformulate rewards and leverage ground-truth answers and combine prior knowledge injection and reflection to optimize model trustworthiness. \methodname{} provides a promising alternative for aligning model preferences with human expectations. Our method achieves substantial performance gains on six video understanding benchmarks, while maintaining minimal training costs. This work not only enhances the alignment but also offers a scalable framework for improving the trustworthiness and effectiveness of Video-LLMs.


\bibliographystyle{unsrt}
\bibliography{neurips_2025}

\appendix

\clearpage

\startcontents[chapters]
\setcounter{section}{0}
\printcontents[chapters]{}{1}{}

\section{Dataset Details}
\label{app:data}
 For a comprehensive evaluation, we conduct our experiment on six several popular video understanding benchmarks, including Video-MME~\cite{fu2024videomme}, LongVideoBench (LongVB)~\cite{wu2024longvideobench}, MLVU~\cite{MLVU}, NeXT-QA~\cite{xiao2021next}, DREAM-1K~\cite{wang2024tarsier} and VideoChatGPT~\cite{maaz2023video}.

 \noindent\textbf{Video-MME} is a large-scale benchmark for evaluating multimodal understanding in video-based LLMs, covering tasks like question answering, captioning, and spatiotemporal reasoning. It includes a diverse set of real-world videos with comprehensive annotations for objective evaluation.

\noindent \textbf{LongVideoBench (LongVB)} focuses on testing the ability of the models to handle extended video content, requiring both fine-grained moment analysis and higher-level summarization. Its annotated videos often span many minutes or even hours, challenging existing context-length limits.

\noindent \textbf{Multi-Level Video Understanding (MLVU)} presents a dataset covering tasks from basic object-action recognition to high-level story comprehension. It tests a model’s capacity for reasoning across semantic levels in diverse real-world video clips.

\noindent \textbf{NeXT-QA (Narrative and Explainable Temporal QA)} emphasizes temporal and causal reasoning in everyday video scenarios. Questions and annotations target multi-step reasoning over sequences of events rather than mere visual recognition.

\noindent \textbf{DREAM-1K} is an open-ended video QA dataset with around 1,000 clips, encouraging free-form, detailed responses. Its tasks promote rich descriptive answers and deeper inference beyond simple factual recall.

\noindent \textbf{VideoChatGPT} is a multi-turn, conversational framework for discussing and reasoning over video content. It tests a system’s capacity to maintain context and coherence across extended visual dialogues.

\section{Visualization of Implicit Reward}
\begin{figure}[]
    \centering
    \includegraphics[width=0.7\linewidth]{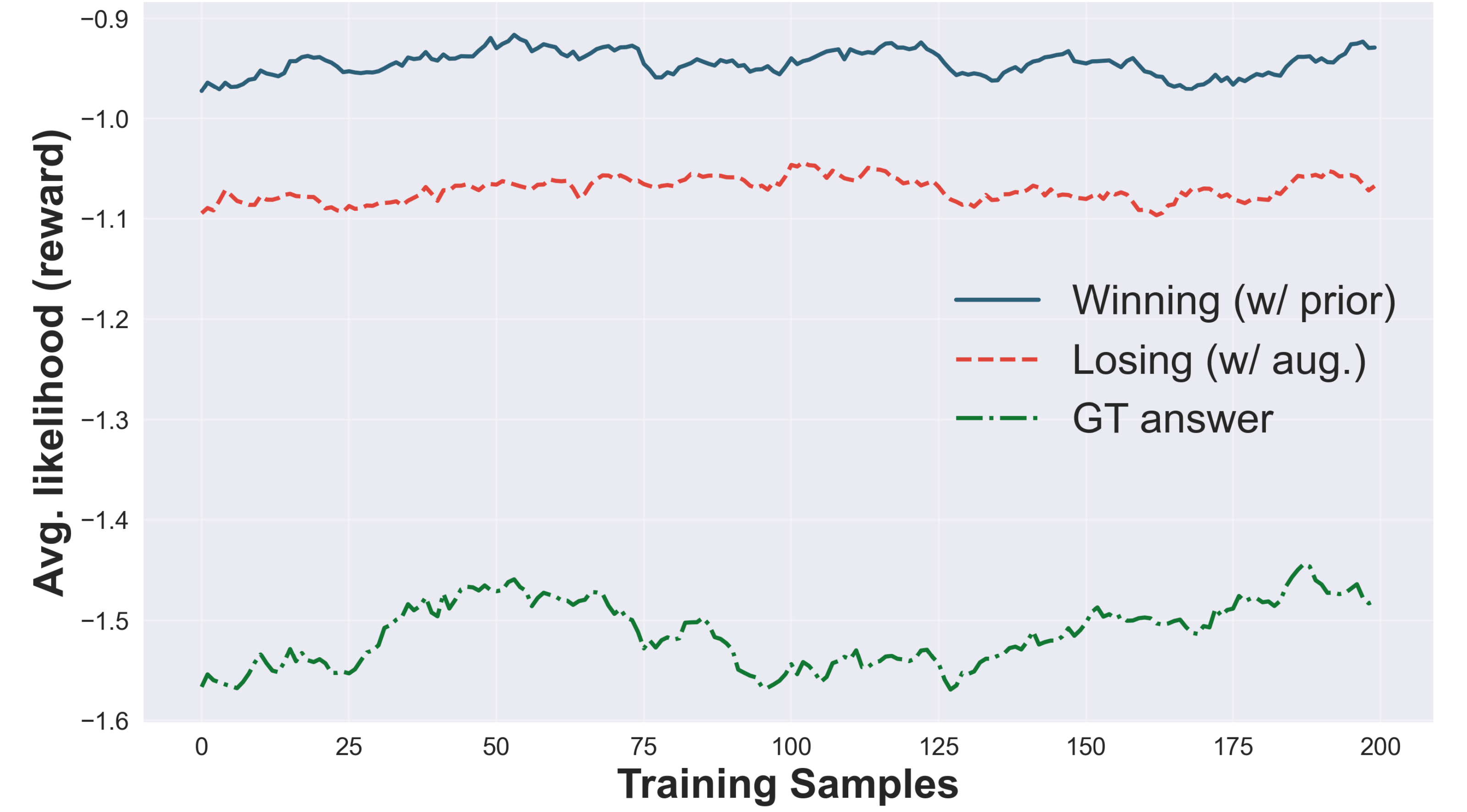}
    \caption{Visualization of average likelihood (implicit reward) for each video sample with different response inputs to the  LLaVA-NeXT-Video-7B model~\cite{zhang2024llavanextvideo}. Samples are randomly chosen from our self-generated preference data.}
    \label{fig:logp}
\end{figure}

To verify the self-generated preference data quality from Sec.~\ref{subsec:response}, as shown in Fig.~\ref{fig:logp}, we visualize the implicit reward for each video sample with different response inputs. These preferred winning responses have both high trustworthiness and high likelihood, which has higher trustworthiness than those responses without prior, and also has higher likelihood than GT answers because the latter does not follow the Video-LLM's output distribution. Although the ground-truth answers are correct for the video queries (and we consider them with the highest trustworthiness), their corresponding likelihood curve is much lower than the other two responses as they fall outside the output distribution of Video-LLMs. Consequently, treating ground-truth answers as equally preferable to the winning responses in preference learning proves inefficient, leading to suboptimal experimental results. This motivated us to develop the above framework to collect winning responses that achieve both high trustworthiness and high likelihood.

\section{Preference Data  Pipeline: DPO vs. LeanPO}
\begin{figure*}
    \centering
    \includegraphics[width=0.95\linewidth]{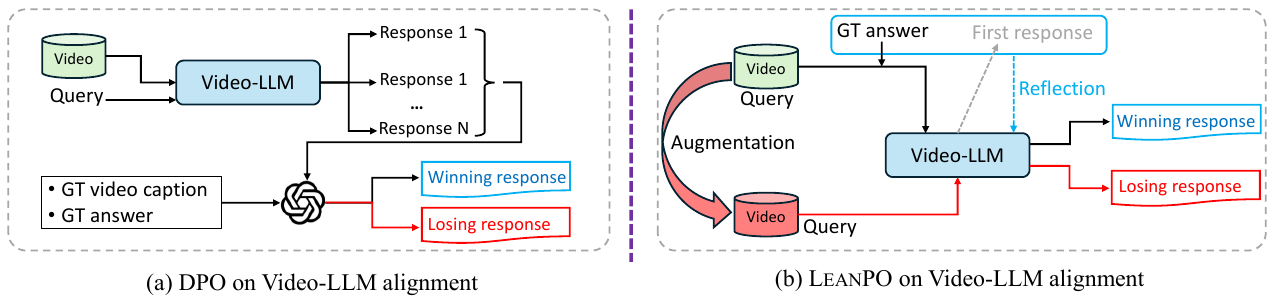}
    \caption{Comparison of different methods for Video-LLM alignment. We compare the preference data generation pipeline between DPO and our proposed \methodname{} on a Video-LLM.}
    \label{fig:data_gen}
\end{figure*}

As shown in Fig~\ref{fig:data_gen}. (a), most existing methods sample multiple responses from Video-LLM itself, and require human or additional strong AI models (such as GPT-4V~\cite{gpt4v}) to label the paired data~\cite{zhang2024hound,ahn2024isrt,vlm-rlaif}, which is expensive and inefficient. To develop a cheaper, effective, and flexible framework to build the own preference data of Video-LLMs, we obey the reward-trustworthiness-correlated principle to construct preference data, as shown in the upper of Fig.~\ref{fig:data_gen}. (b).

\section{More Case Analysis}
\begin{figure}
    \centering
    \includegraphics[width=0.7\linewidth]{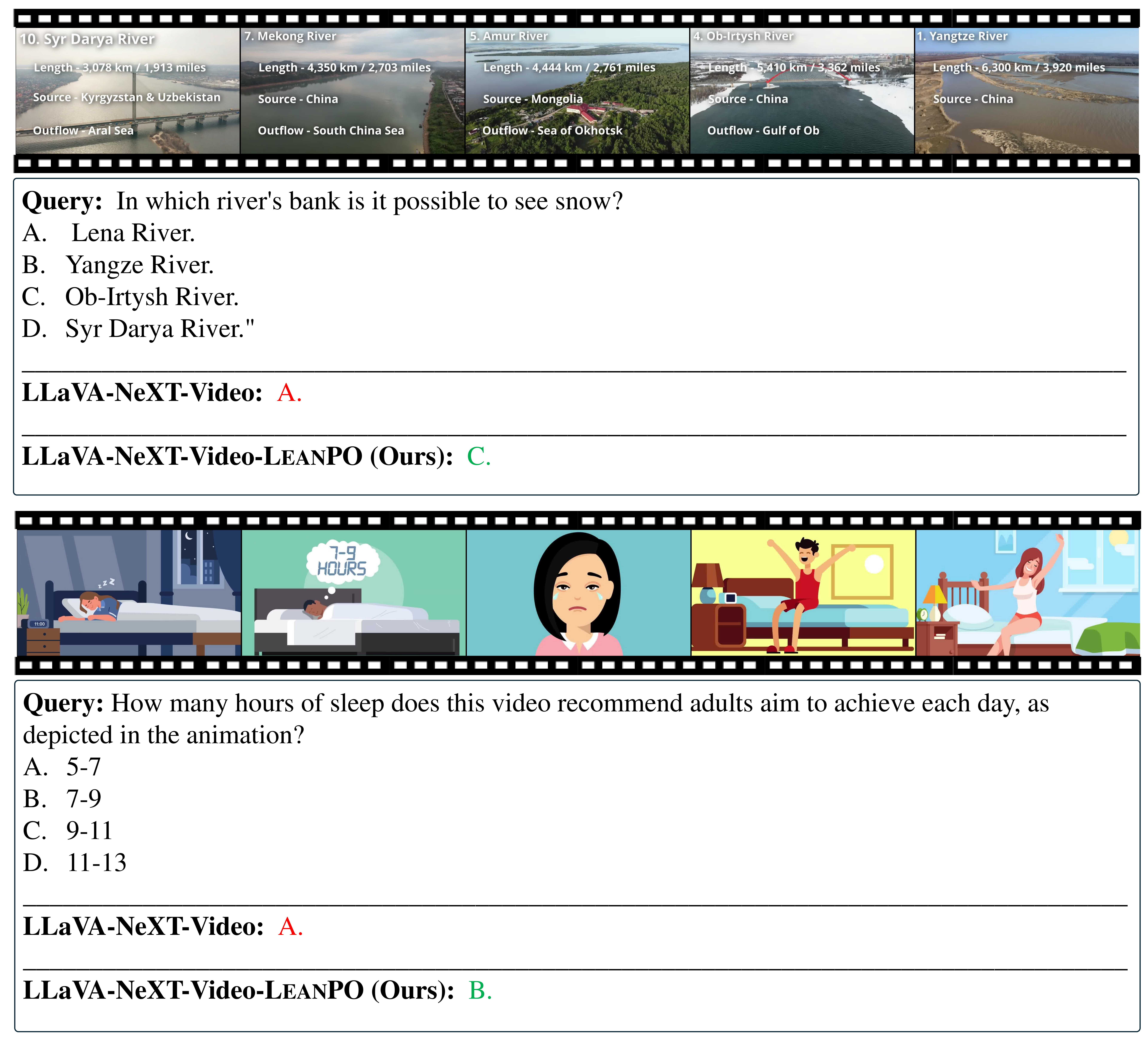}
    \caption{More Video QA results compared with baseline.}
    \label{fig:moreqa1}
\end{figure}

\begin{figure}
    \centering
    \includegraphics[width=\linewidth]{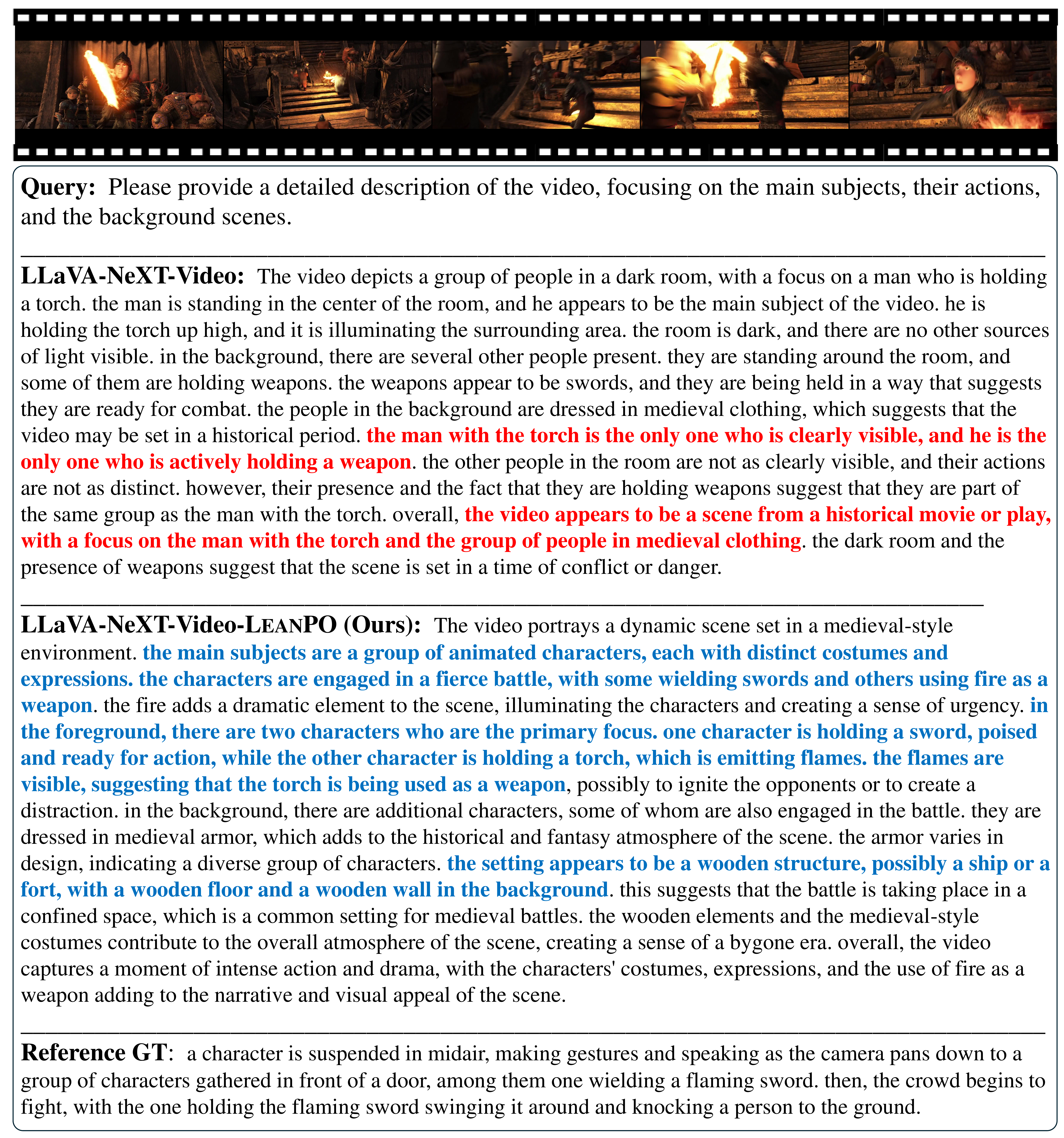}
    \caption{More Video detailed description results compared with baseline.}
    \label{fig:moredc1}
\end{figure}

\begin{figure}
    \centering
    \includegraphics[width=\linewidth]{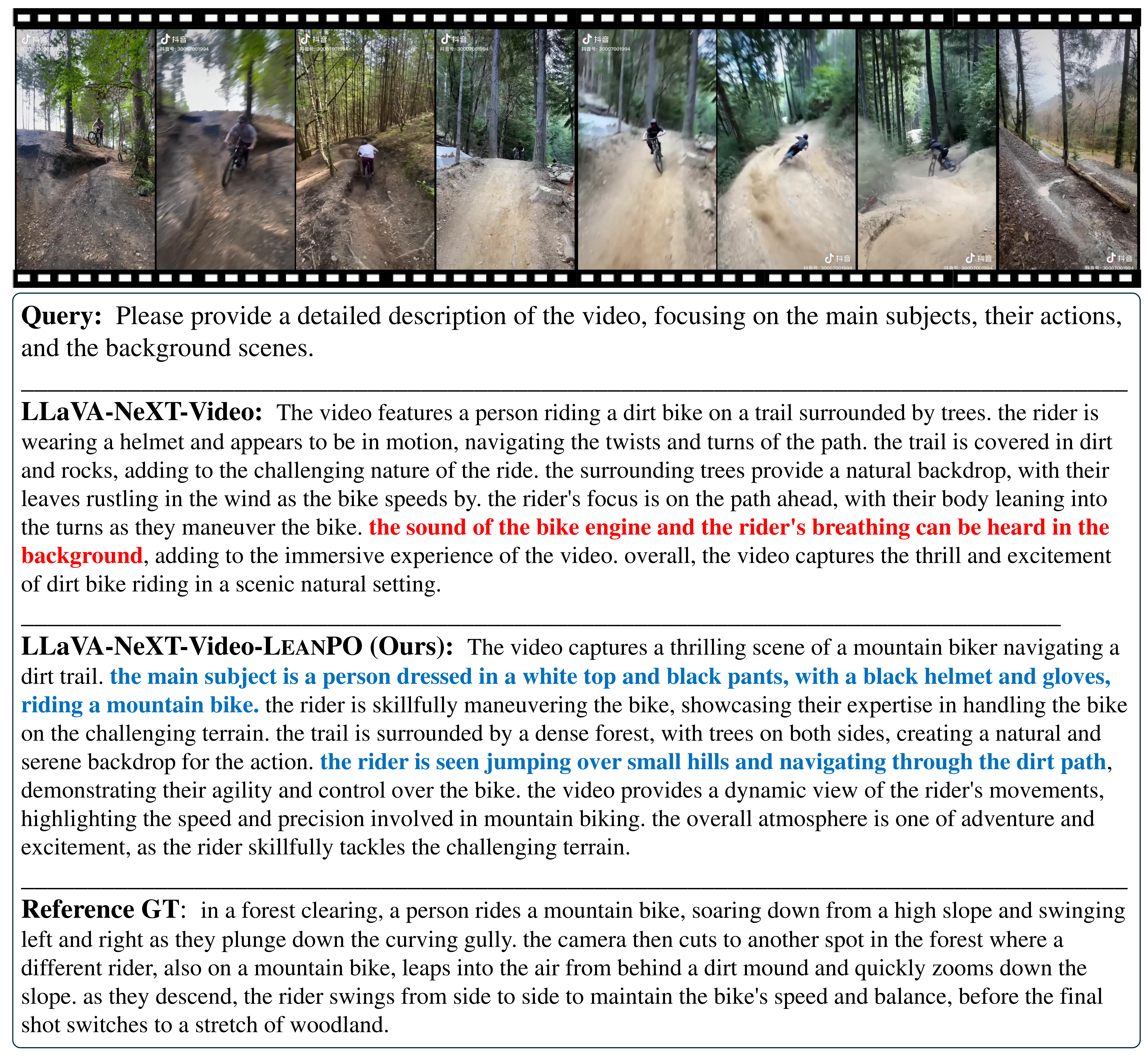}
    \caption{More Video detailed description results compared with baseline.}
    \label{fig:moredc2}
\end{figure}

We present more qualitative results for Video Multiple-choice QA and video detailed description in Fig.~\ref{fig:moreqa1}, Fig.~\ref{fig:moredc1}, Fig.~\ref{fig:moredc2}.


\end{document}